\def\year{2021}\relax
\documentclass[letterpaper]{article} 
\usepackage{aaai21}  
\usepackage{times}  
\usepackage{helvet} 
\usepackage{courier}  
\usepackage[hyphens]{url}  
\usepackage{amsmath}
\usepackage{amssymb}
\usepackage{bm}
\usepackage{subfigure}
\usepackage{graphicx} 
\usepackage{natbib}  
\usepackage{caption} 
\title{Semi-Structured Deep Piecewise Exponential Models}
\author{
    Philipp Kopper\textsuperscript{\rm 1},  Sebastian P{\"o}lsterl\textsuperscript{\rm 2},  Christian Wachinger\textsuperscript{\rm 2}, Bernd Bischl\textsuperscript{\rm 1},
     Andreas Bender\textsuperscript{\rm 1,}\footnote{Contributed equally.}, 
     David R{\"u}gamer\textsuperscript{\rm 1,$\ast$} 
    \\
}
\affiliations{
    \textsuperscript{\rm 1} Chair of Statistical Learning and Data Science, Department of Statistics, LMU Munich, Germany\\
        \textsuperscript{\rm 2} Artificial Intelligence in Medical Imaging, Department of Child and Adolescent Psychiatry, LMU Munich, Germany\\
             \{philipp.kopper, bernd.bischl, andreas.bender, david.ruegamer\}@stat.uni-muenchen.de,
        \{sebastian.poelsterl, christian.wachinger\}@med.uni-muenchen.de\\

}

\begin{document}

\maketitle

\begin{abstract}
We propose a versatile framework for survival analysis that combines advanced concepts from statistics with deep learning. The presented framework is based on piecewise exponential models and thereby supports various survival tasks, such as competing risks and multi-state modeling, and further allows for estimation of time-varying effects and time-varying features.  To also include multiple data sources and higher-order interaction effects into the model, we embed the model class in a neural network and thereby enable the simultaneous estimation of both inherently interpretable structured regression inputs as well as deep neural network components which can potentially process additional unstructured data sources. A proof of concept is provided by using the framework to predict Alzheimer's disease progression based on tabular and 3D point cloud data and applying it to synthetic data.
\end{abstract}

\section{Introduction}

Similar to other fields, survival analysis has greatly benefited from the influx of machine learning (ML) and deep learning (DL) methods. However, especially early adaptations only focused on improved predictions of the event times or the survival probability in standard settings, i.e., right-censored, single-event data.

In this paper, we introduce a new method for continuous time-to-event data that combines advanced concepts from statistics and deep learning to form a new, versatile framework for survival analysis. The proposed framework enables the hazard-based learning of survival models via neural networks, which supports 1) survival tasks with right-censored, left-truncated, competing risks, or multi-state data; 2) estimation of inherently interpretable feature effects; 3) learning from multiple data sources (e.g., tabular and imaging data);
4) time-varying effects and time-varying features;
5)  modeling of functional features (e.g., cumulative effects). 

The new approach combines two general frameworks, the so-called piecewise exponential additive model (PAM) and semi-structured deep distributional regression (SDDR). The former provides a general framework for survival analysis that supports estimation of interpretable, additive effects of tabular features (referred to as the \emph{structured} predictor in the following) and the latter embeds this structured part in a neural network (NN) while simultaneously explaining remaining heterogeneity or incorporating unstructured data (e.g., images or text) using additional NN components. Similar to the Cox model, our model does not require assumptions regarding the distribution of event times.

Before introducing the framework more formally in Section~\ref{sec:PAM} and \ref{sec:deeppam}, we briefly recap the relevant literature and developments in relation to PAMs and neural network-based approaches to survival analysis.

\paragraph{Piecewise Exponential Additive Models} The piecewise exponential additive model or PAM is an extension of the piecewise exponential model (PEM).  The original formulation of the PEM, a parametric, linear effects, proportional hazards (PH) model, goes back to \citet{Holford1980, Laird1981, Friedman1982}. The general idea is to partition the follow-up time into $J$ intervals $(\kappa_{j-1},\kappa_j],\ j=1,\ldots,J$, and to assume piecewise constant hazards in each interval. Although the PEM was shown to be (mostly) equivalent to the Cox PH model \citep{Whitehead1980}, the latter prevailed in the survival literature, as the originally proposed PEM requires a careful choice of the number and placement of interval cut-points in a trade-off between the flexible estimation of the baseline hazards (large $J$) on the one hand and robustness of the estimates on the other hand (number of events per interval decreases as $J$ increases). 

Following \citet{Cai2002, kauermann_penalized_2005, argyropoulos_analysis_2015, Bender2018b}, PAMs estimate the baseline hazard and other time-dependent effects via penalized splines, which leads to smooth and more robust estimates of the baseline hazards. Furthermore, the number of parameters only depends on the basis dimension of the spline, rather than the number of intervals $J$. This makes PAMs computationally more efficient in high-dimensional settings \citep{greven_comments_2020}.

\paragraph{Neural Network-based Approaches} Neural network-based approaches for survival analysis also received a lot of attention. Early use of NNs for Cox type models was proposed by \citet{faraggi_neural_1995}. More recently, various DL-based approaches have been proposed, most prominently: a single event survival model using deep exponential families \citep{ranganath_deep_2016}, a framework based on Gaussian processes for competing risks data \citep{alaa_2017}, and DeepHit a framework for discrete time-to-event data which can handle competing risks \citep{lee_deephit_2018}, which was recently extended to handle time-varying features \citep{lee_dynamic-deephit_2020}.

The first combination of PEMs with a NN was proposed by \citet{liestbl_survival_1994}. \citet{biganzoli_general_2002} discussed the estimation of PEM by representing generalized linear models via feed-forward NNs, and \citet{fornili_piecewise_2014} proposed the estimation of the shape of the hazard rate with NNs, rather than parametrically. \citet{kvamme_continuous_2019} also discussed the parametrization of the PEM via NNs with application to tabular data. Similar to the discussion above for structured PEMs, they found that the choice of cut-points is crucial for performance. As an extension of PAMs, the framework proposed in the following sections largely eliminates this problem.\\

\noindent The remainder of the paper is organized as follows:
In Section~\ref{sec:PAM} we will formally introduce the PAM, followed by its combination with SDDR in Section~\ref{sec:deeppam}. As a proof of concept, Section~\ref{sec:numexp} illustrates the application of the new framework to tabular and point cloud data with the goal to predict Alzheimer's disease progression and to synthetic data.

\section{Piecewise Exponential Additive Models} \label{sec:PAM}

\subsection{General Model Definition}\label{ssec:PAM_definition}
The general model specification of PAMs is given by
\begin{equation}\label{eq:general_hazard}
h(t|\mathbf{x}(t),k) = \exp\left(f(\mathbf{x}(t),t,k)\right), k=1,\ldots,K
\end{equation}
which defines the hazard $h$ for time point $t\in \mathcal{T}$, conditional on a potentially time-varying feature vector $\mathbf{x}(t) \in \mathbb{R}^{P}$. The function $f(\cdot)$ represents the effect of (time-dependent) features $\mathbf{x}(t)$ on the hazard and can itself be potentially time- and transition-specific. $k$ indicates a transition, e.g., from status $0$ to status $k$ in competing risks, or the transition between two states in the multi-state setting.
We will omit the dependence on $k$ for readability in the following and focus on survival tasks without competing risks or multiple states. Further omitting the dependence on $t$, \eqref{eq:general_hazard} reduces to the familiar PH form known from the Cox model.

\subsection{Data Transformation}\label{ssec:ped}
PEMs and PAMs approximate \eqref{eq:general_hazard} via piecewise constant hazards, which requires a specific data transformation, creating one row in the data set for each interval a subject was at risk. Assume $n$ observations (subjects), $i=1,\ldots,n$, for which the tuple $(t_i,\delta_i,\bm{x}_i)$ with event time $t_i$, event indicator $\delta_i \in \{0,1\}$ (1=event, 0=censoring) and feature vector $\bm{x}_i$ being observed. PAMs partition the follow up into $J$ intervals $(\kappa_{j-1},\kappa_j],\ j=1,\ldots,J$.  
This implies a new status variable $\delta_{ij} = 1$ if $t_{i} \in (\kappa_{j-1}, \kappa_j] \wedge \delta_{i} = 1$, and $0$ otherwise, indicating the status of subject $i$ in interval $j$. Further, we create a variable $t_{ij}$, the time subject $i$ was at risk in interval $j$, which will enter the analysis as an offset. Lastly, the variable $t_j$, (e.g., $t_j:=\kappa_j$) is a representation of time in interval $j$ and the feature based on which the model estimates the baseline hazard and time-varying effects. To transform data into the piecewise exponential data format (PED), time-constant covariates $\bm{x}_{i}$ are repeated for each of $J_i$ rows, where $J_i$, denotes the number of intervals in which subject $i$ was at risk. This data augmentation step transforms a survival task to a standard Poisson regression task. Depending on the censoring type, e.g., right-censoring, competing risks, or left truncation, the specifics of the data transformation vary, but the general principles remain the same. For more details refer to \citet{pammtools, bender2020general}.

\subsection{Model Estimation}\label{ssec:pam_estimation}
Given the transformed data, PAMs approximate \eqref{eq:general_hazard} by
\begin{equation}
    h(t|\mathbf{x}_i(t)) = \exp(f(\mathbf{x}_{ij},t_j)):=h_{ij}, \forall t \in (\kappa_{j-1},\kappa_j],
\end{equation}
where $\mathbf{x}_{ij}$ is the feature vector of subject $i$ in interval $j$.

Assuming $\delta_{ij}\sim \mathrm{Poisson}(\mu_{ij}=h_{ij}t_{ij})$, the log-likelihood contribution of subject $i$ is given by 
\begin{equation} \label{eq:lik-contrib}
\ell_i = \sum_{j=1}^{J_i}(\delta_{ij}\log(h_{ij}) - h_{ij}t_{ij}),
\end{equation}

Equation \eqref{eq:pam_eta} shows one standard parametrization of $h_{ij}$ in the context of PAMs: 
\begin{equation}\label{eq:pam_eta}
\log(h_{ij}) = \beta_0 + f_0(t_j) + \sum_{p=1}^P x_{ij,p} \beta_p + \sum_{l=1}^L f_l(x_{ij,l}),
\end{equation}
with log-baseline hazard $\beta_0 + f_0(t_j)$, linear feature effects $\beta_p$ and univariate, non-linear feature effects $f_l(x_{ij,l})$ of features $x_{ij,p},x_{ij,l}\in\bm{x}_{ij}$.
Both $f_0$ and $f_l$ are defined via a basis representation, i.e., 
$$f_l(x_{ij,l}) = \sum_{m=1}^{M_l}\theta_{l,m}B_{l,m}(x_{ij,l})$$ with basis functions $B_{\cdot,m}(\cdot)$ (such as B-spline bases) and basis coefficients $\theta_{\cdot,m}$. To avoid underfitting, the basis dimensions $M_0$ (for $f_0$) and $M_l$ (for $f_l$) are set relatively high. To avoid overfitting, the basis coefficients are estimated by optimizing an objective function, that penalizes differences between neighboring coefficients. Let $\boldsymbol{\beta} = (\beta_0,\ldots,\beta_P)^\top$ and $\boldsymbol{\theta}_{l}=(\theta_{l,1},\ldots, \theta_{l,M_l})^\top$, $l=0,\ldots,L$. The objective function minimized to estimate PAMs is the penalized negative log-likelihood, which is given by
\begin{equation}\label{eq:penalized_likelihood}
- \log \mathcal{L}(\boldsymbol{\beta},\boldsymbol{\theta}_0,\ldots,\boldsymbol{\theta}_L) + \sum_{l=0}^{L} \psi_{l}J(\boldsymbol{\theta}_{l}),
\end{equation}
where the first term is the standard negative logarithmic Poisson likelihood, comprised of likelihood contributions \eqref{eq:lik-contrib}, and the second term $J(\boldsymbol{\theta}_{l})$ is a quadratic penalty with smoothing parameter $\psi_{l}\geq 0$ for the respective spline $f_l$. Larger $\psi_{l}$ lead to smoother $f_l$ estimates \citep[see][for details]{wood_generalized_2017, Bender2018a}. 

In the following, we will subsume all coefficients of structured additive model components such as $\boldsymbol{\beta}$ or $\boldsymbol{\theta}_l$ in the vector $\bm{w}$. All structured features can be furthermore summarized in a design matrix $\bm{B}$. This means we can represent the hazard as $\log(h_{ij}) = \bm{B}_{ij}\bm{w}$.

\section{Semi-structured Learning with DeepPAM} \label{sec:deeppam}

DeepPAMs extend PAMs as defined in \eqref{eq:general_hazard} by including unstructured model inputs into the additive predictor $f(\mathbf{x}(t),t)$. While PAMs restrict $f$ to structured additive effects, the hypothesis space of DeepPAMs can also include $G$ (deep) NN components $d_g(\cdot), g=1,\ldots,G$. For readability, we assume only one NN component $d(\cdot)$ in the following. This NN predictor is used to process a potentially time-varying unstructured data source $\bm{z}(t)$. Different ways exist to integrate $d(\cdot)$ into PAMs, depending on the assumed relationship of structured and unstructured data. Here, we distinguish between two types of model classes, derived from the role of $d$. The first class of DeepPAMs assumes a time-constant effect of unstructured data sources
\begin{equation}
    \label{eq:deeppem_1}
    h(t|\bm{x}(t),\bm{z}(t)) = \exp\bigl\{ f(\mathbf{x}(t),t) + d(\bm{z}(t))\bigr\},
\end{equation} 
which enriches the structured predictor of PAMs using one (or in general $G$) NN predictor(s). 
The second, more flexible version of DeepPAMs allows for an interaction between structured and unstructured predictors:
$$h(t|\bm{x}(t),\bm{z}(t)) = \exp\bigl\{f\left(\mathbf{x}(t),d(\bm{z}(t)),t\right)\bigr\},$$ where $f$ now also depends on the unstructured data sources and the specified NN(s). As in standard PAMs, dependence on $t$ allows deviations from the PH assumption. 

Instead of combining PAMs with (deep) neural networks in a two-stage approach, we embed PAMs in a neural network using the idea of SDDR and train the network based on the (penalized) likelihood defined in \eqref{eq:penalized_likelihood}. A general framework for end-to-end learning of \emph{structured} (additive) predictors and neural network components has recently been proposed in \citet{rugamer_semistructured_2021} and successfully applied in the context of transformation \cite{baumann2020deep} and mixture \cite{ruegamer2020neural} models.

\subsection{PED and Latent Representations}

\begin{figure*}
    \centering
    \includegraphics[width = 0.95\textwidth]{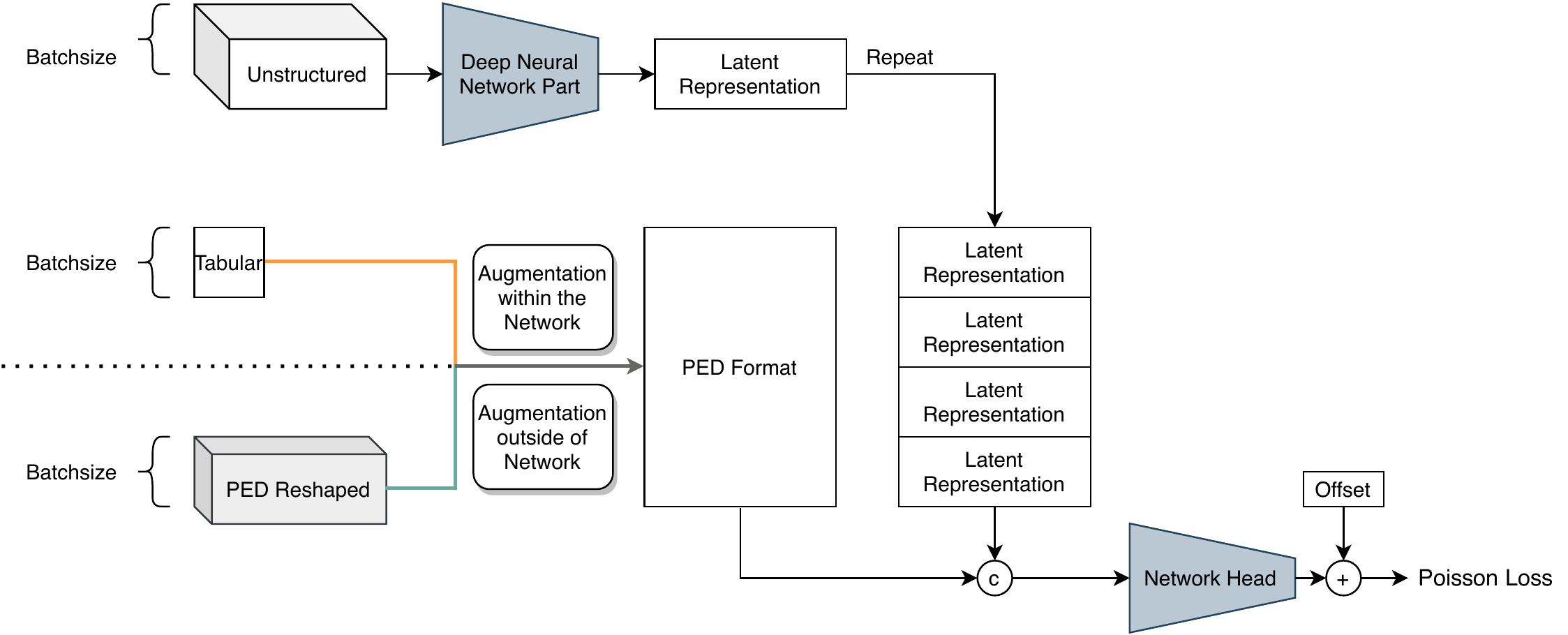}
    \caption{Architecture of a DeepPAM. The unstructured data, e.g., the images, are summarized to latent representations and combined with the tabular data in PED format using a concatenation (upper and right part of the image). The tabular data is either fed into the network in raw format and augmented within the network (orange path) or augmented outside of the network prior to the network training and reshaped to fit the defined batch size (green path). Latent representations of unstructured predictors are learned in a separate deep NN part and concatenated (c) with the transformed structured predictors by repeating the former. The joint influence of the latent representations and structured inputs is learned using one (or more) additional fully-connected layer(s) on top of the concatenated features (Network Head). Finally, the offset is added to the prediction and the network is trained using the Poisson loss. }
    \label{fig:deepPAM}
\end{figure*}

As outlined in section \ref{ssec:ped}, the application of the framework requires a data transformation step. This brings many advantages, but one drawback is the increased data size. Thus, a major challenge when combining PED formatted data with unstructured data sources lies in the size and computational costs associated with each datum. The augmentation of tabular data can usually be done in memory. However, repeating unstructured data such as images for each interval of each observation requires multiple forward-passes of identical data through the network. This results not only in notably longer runtimes, but may also be redundant if the unstructured data sources are constant over time. For example, if we only have access to a patient's brain scan at the start of the study, it will not change for the remaining time intervals.

We address this issue in DeepPAMs using one of two options. Figure~\ref{fig:deepPAM} schematically illustrates the two data handling pipelines within the network. 
Option 1 allows for PED format augmentation outside of the network. Instead of feeding the original tabular data into the network, the PED format is created prior to the network training and reshaped PED tensor batches with the same sampling dimension as the unstructured data source are passed through the network and then flattened internally to recover the original PED format. This option is especially useful when the network cannot deal with varying batch dimensions internally or the pre-processing has to be done outside the network for other reasons. The second option transforms the original tabular data into the PED format within the network. This can be more flexible as the PED format does not need to be transformed into a (ragged) tensor or must be padded as different subjects are usually observed for a varying number of intervals. 

In both cases, we avoid repeating the original unstructured data source multiple times while still able to learn a common latent representation for the unstructured source for all time points of each subject. We achieve this by repeating the latent representation that is learned from the unstructured data source in the network. Finally, we combine these representations with the original tabular data. 

In the following section, we will present a concrete architecture implementing this process together with the underlying Poisson loss optimization problem to learn a piecewise constant hazard from unstructured sources.

\subsection{Learning piecewise Constant Hazards from Unstructured Data Sources} \label{sec:deeppam_linear}

To make our concept of DeepPAM more concrete, we will now describe how the framework can be used to learn time-constant latent representations through an additive deep neural network predictor. This resembles the data generating process of the data we will later apply the framework to.

For simplicity assume a $Q$-dimensional unstructured data source $\bm{z}_i \in \mathbb{R}^{q_1 \times \ldots \times q_Q}$, e.g. an greyscale image with height $q_1$ and width $q_2$. $\bm{z}_i$ is passed through DeepPAM along with tabular inputs $\bm{x}_i$. If we assume that $\bm{z}_i$ has a piecewise constant effect on the hazard, i.e., an multiplicative effect on the baseline hazard, the corresponding DeepPAM is defined as in \eqref{eq:deeppem_1}. While feeding the data into the network in batches defined on the subject level $i=1,\ldots,n$, the resulting additive predictor on the level of the PED is given as follows:
$$h_{ij} = \exp\bigl\{\bm{B}_{ij}\bm{w} + \sum_{u=1}^U \zeta_{ij,u} \gamma_u \bigr\}.$$
Here, $\zeta_{ij,1},\ldots,\zeta_{ij,U}$ are $U$ latent representations learned from the unstructured data source $\bm{z}_{i}$ and $\gamma_1,\ldots,\gamma_U$ the corresponding feature effects. The $\gamma$s are jointly learned with the weights $\bm{w}$ of the structured network part in a shared network head, which is a 1-unit hidden layer with linear activation. The corresponding architecture is visualized in Figure~\ref{fig:deepPAM}.

\section{Numerical Experiments} \label{sec:numexp}

In the following, we apply the proposed method using a real-world data set on Alzheimer's Disease (Section~\ref{sec:application}) and a synthetic data set (Section~\ref{sec:simulation}) in order to study the behavior of our model.

\subsection{Learning the Risk of Alzheimer's Disease Progression} \label{sec:application}

Alzheimer's disease (AD) is mostly diagnosed among elderly individuals from their 60s on.
Early stages imply mild cognitive impairment (MCI).
Patients suffering from MCI are not impaired in their daily life while they still face significant cognitive symptoms.
Not all patients suffering from MCI eventually develop dementia symptoms \citep{petersen2011mild, langa2014diagnosis}.
In our experiments, we use the data from the Alzheimer’s Disease Neuroimaging Initiative \citep[ADNI;][]{jack2008alzheimer} to predict the progression of patients with MCI to dementia.
Since 2003 ADNI has collected data of different modalities such as magnetic resonance imaging (MRI) and positron emission tomography (PET) and other biomarkers, clinical and neuropsychological assessments.

Using data from ADNI, our main goal is to study whether medical images have additional predictive capacity when also accounting for commonly used predictors such as non-image biomarkers.

\paragraph{Data}

We use a subset of the ADNI database from \citet{polsterl2019wide}, including 397 individuals with MCI at their baseline assessment.
MRI scans of the patient were processed using FreeSurfer \citep{fischl2012freesurfer} and ShapeWorks \citep{cates2008particle} to eventually obtain smooth left hippocampi surfaces represented as point clouds.
We downsample the point clouds to 1,024 coordinates, normalize the coordinates as proposed by \citet{qi2017pointnet} and add noise on the point clouds used for generalizability \citep[cf.][]{qi2017pointnet}.
Next to the medical images, we make also use of tabular clinical data including the level of education (years of education binned to 4 intervals), age, and sex.
Furthermore, we account for relevant predictive biomarkers for the progression of AD:
FDG-PET, AV45-PET, levels of beta-amyloid 42 peptides (Abeta), total tau protein (Tau).

As an additional feature derived from the images we use the left hippocampus volume.
Literature finds a link between structural changes in the hippocampus and proceeding dementia \citep{wachinger2016whole, gerardin2009multidimensional, frisoni2008mapping}.
This motivates the use of hippocampi data in the analysis.

\paragraph{Model}

For the unstructured deep part of the model, we use a variation of the PointNet \citep{qi2017pointnet} with a reduced number of weights similar to the parametrization suggested in \citet{polsterl2019wide}. 
Our PointNet comprises a series of shared multi-layer-perceptrons (MLP), defined as 1D convolutions with kernel size 1, with batch normalization and ReLu-activation, global max pooling , and a series of global MLPs.
We use the PointNet to learn 400 latent global features of the left hippocampi shape, which we further process through MLPs to one latent representation. 
Details on the architecture can be found in \citet{polsterl2019wide}.

For the structured model part, we use linear effects for biomarkers and the hippocampus volume, a one-hot encoded linear feature effect for education and a smooth B-spline representation of age with five equidistant knots. Finally, we use a smooth B-spline representation of the time in order to model the baseline hazard. For spline regularization, we use a quadratic penalty term with its smoothing parameter ($\psi_0$) taken from the baseline PAM (cf. Section \ref{ssec:pam_estimation} and Equation \eqref{eq:penalized_likelihood}). We use the PAM estimates for a warm start of the structured part of the network and optimize the model using Adam \citep{kingma2015adam} with learning rate of 0.001.
We perform early stopping based on the validation data performance.

\paragraph{Evaluation} 

We learn the underlying association of clinical features and point clouds with patients' risk to AD progression on 10 different train-validation-test splits.
In each repetition, we use 319 observations for training, 39 for validation, and 39 for testing.
We evaluate our model by comparing the results to a baseline PAM using the same information as in the structured part of our network and also compare our model to the wide-and-deep approach by \citet{polsterl2019wide}. 
Additionally, we include a Cox PH model with the same specification as the baseline PAM.
We use the integrated Brier Score \citep[IBS;][]{graf1999assessment} at the 25\%, 50\%, and 75\% quartiles of the training data as our evaluation criterion to measure predictive performance. 

\paragraph{Results}

We find that all approaches yield comparable predictive performance, although the DeepPAM appears to perform slightly better when evaluated at later time-points compared to W\&D.
In Table~\ref{tab:res} none of the models outperforms the others substantially.
In particular, neither of the neural network-based approaches are able to improve survival probability estimates by additionally taking medical images in the form of left hippocampi point clouds into account.
Different architectures of the PointNet did not alter these findings notably.\\

\begin{table}[!htbp] \centering 
\begin{tabular}{@{\extracolsep{5pt}} cccc} 
\\[-1.8ex]\hline 
\hline \\[-1.8ex] 
 & 1st Quartile & Median & 3rd Quartile \\ 
\hline \\[-1.8ex] 
KM & $\mathbf{3.0}$ $(2.0)$ & $7.3$ $(3.8)$ & $10.3$ $(4.5)$ \\ 
PAM & $\mathbf{3.0} (2.1)$ & $\mathbf{6.3}$ $(3.4)$ & $7.8$ $(3.4)$ \\ 
DeepPAM & $3.1$ $(2.1)$ & $6.4$ $(3.2)$ & $7.8$ $(3.2)$ \\ 
Cox PH & $\mathbf{3.0}$ $(2.1)$ & $\mathbf{6.3}$ $(3.4)$ & $\mathbf{7.7}$ $(3.4)$ \\ 
W\&D & $\mathbf{3.0}$ $(2.0)$ & $6.8$ $(3.5)$ & $8.6$ $(3.6)$ \\ 
\hline \\[-1.8ex] 
\end{tabular} 
 \caption{Averaged IBS (standard deviation in brackets) for the Kaplan-Meier (KM) estimator, PAM, DeepPAM, Cox-PH and \citet{polsterl2019wide} (W\&D). Kaplan-Meier, Cox PH and PAM do not make use of the point clouds.
 The values have been multiplied by 100 for better readability.}
 \label{tab:res} 
\end{table} 

\noindent In the following, we examine our model in a second, more controlled setup, which allows examining whether the previous null result is an artifact of the proposed model.

\subsection{Synthetic Data} \label{sec:simulation}

We generate synthetic data on the basis of the ModelNet10 data set \citep{wu20153d}, also containing point clouds.
We enrich this data set by simulating additional tabular input features and finally simulate right-censored survival times.

\paragraph{Data and Simulation}

The ModelNet10 data set consists of point clouds of ten different items. We first reduce the initial data set to the first three items (classes 0, 1, and 2) to allow for a better visual inspection of the model. Each item is stored as a set of coordinates in three dimensions (the x-, y- and z-coordinate) and can have different lengths. 
To reduce computational costs without losing too much information, we downsample the data of each point cloud to 1,024 points.
We add a small uniformly distributed error on each coordinate for training. We further normalize each point cloud as proposed by \citet{qi2017pointnet}. The final synthetic data set consists of 1,008 observations for training, 144 observations for validation, and 216 for testing.

In order to relate the point clouds to a simulated survival time, we treat the item label as one-hot encoded categorical features.
We assign the dummies different linear effects on the log hazard. 
We further simulate two additional uniformly distributed features $x_1, x_2$. These features are defined to have a linear effect on the log hazard. Finally, we simulate a time-varying baseline hazard. 

The hazard, based on which the survival times are simulated is thus given as follows:

\begin{equation}
\begin{split}
h_{i}(t) &=\exp(\beta_0 + f_{0}(t) + \beta_1 x_{i,1} + \beta_2 x_{i,2} +\\
    & \gamma_1 \Delta_{i,1} + \gamma_2 \Delta_{i,2}),
    \end{split}
\end{equation}
where the log baseline hazard is $\beta_0 + f_{0}(t) = -0.5 - 0.1  \cdot (t - 4) ^ 2$ with coefficients $\beta_1 = - 0.25$ and $\beta_2 = 0.3$.
$\Delta_1$ and $\Delta_2$ are dummy-encoded representations of the indexed item.
The linear item effects $\gamma_1 = 0.5$, $\gamma_2 = -1$. 
All survival times are administratively censored at $t = 10$. Additionally, we introduce right-censoring.
The censoring time follows an exponential distribution with a rate of $\lambda = 0.02$.

The challenging part of the synthetic data set is that the latent class information, $\Delta_i, i=0,1,2$, is unknown to the model and can only be approximated through a linear effect of the learned point cloud representation. 
Thus, the defined model does not perfectly mimic the true data generating process but estimates a surrogate for the classes. 

\paragraph{Model}

We model the data generating process using the proposed DeepPAM as described in Section~\ref{sec:deeppam_linear}.
The structured model part is defined in accordance with the true data generating process. For the point clouds we use a similar architecture as described in the previous subsection, but with different numbers of units in each layer. 
We reduce the number of parameters in the unstructured part of the network to 51,585 in total and use L2 regularization instead of dropout for regularization.
The model is trained for a maximum of 75 epochs using Adam with a learning rate of 0.001. We employ early stopping based on the model's performance on the validation set. 

\paragraph{Evaluation}

For training, validation, and testing, we split the data into three parts with approximately 75\%, 10\%, and 15\%, respectively. We repeat our experiments 10 times using different simulated data. We evaluate effect estimates learned from point clouds using visual inspection  and the model's predictive performance on a held-out test data set using the IBS. 

We compare DeepPAM to a PAM model that is specified correctly w.r.t. to the data generating process (correct PAM) as well as a baseline PAM model, that does not include any point cloud information. The baseline model is used to check the model performance when no information from the point cloud items is taken into account and is expected to just average over the simulated linear item effects. The correct PAM is supplied with the (in reality) unknown correct label of the respective point cloud and should thus be able to recover the item-specific hazards.

\begin{figure}[htp]
    \centering
    \includegraphics[width=8cm]{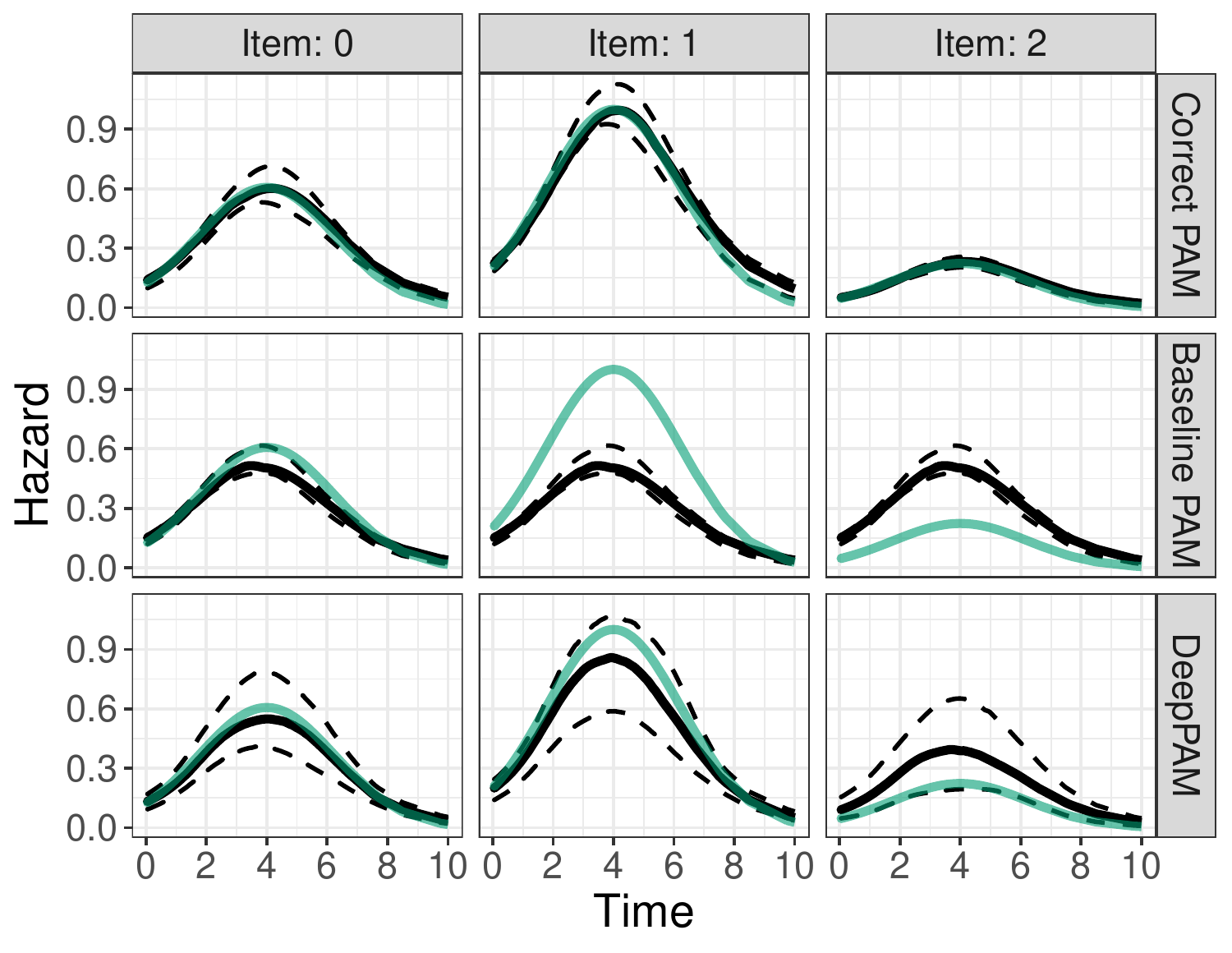}
    \caption{Median (black) and empirical 5\%- and 95\%-quantiles (dashed) of estimated hazard effects of DeepPAM. Effects are compared to the true effect (cyan) for each of three point cloud labels (columns) and each model (rows).}
    \label{fig:simul}
\end{figure}

\paragraph{Results}

\begin{figure}[htp]
    \centering
    \includegraphics[width=8cm]{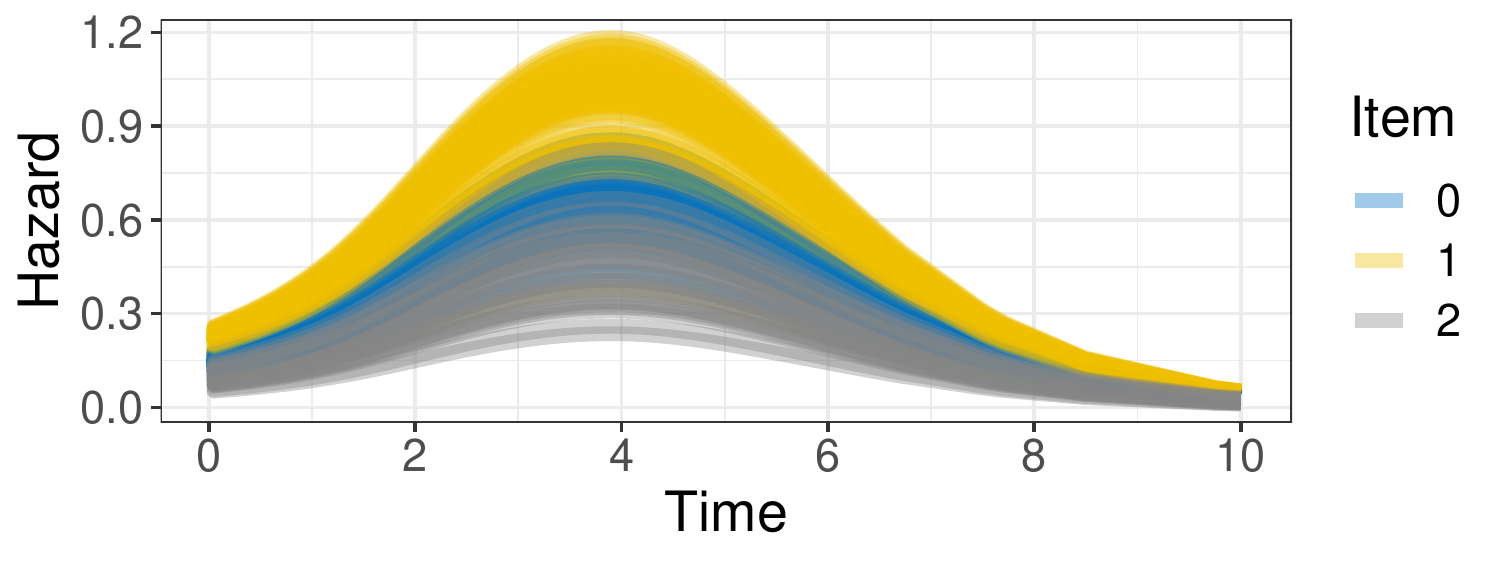}
    \caption{An example of the estimated baseline hazard effects by DeepPAM for different items (colors).}
    \label{fig:simul-overlap}
\end{figure}

Figure~\ref{fig:simul} shows the resulting estimates obtained from DeepPAM. Every single model plot corresponds to the distribution of 2,160 predicted hazard values for each latent item class, characterized by their median, their 5\%- and 95\%-quantile together with  the true effect. As a comparison, we also plot the predictions of both PAMs.
Figure~\ref{fig:simul-overlap} additionally displays the complete distribution of predicted latent effects for a single simulation run for all different items.

Results indicate that the baseline model without additional class information predicts a constant hazard for all three classes. In contrast, DeepPAM is able to use the additional unstructured information and better discriminates between the three classes. 
The model recovers the effects of the two first classes very well. 
Although the effect of the third class is biased towards the mean effect, the true effect lies within the plotted quantile range.
This means that our model behaves in the predicted behavior.
All models tend to recover the linear coefficients $\beta_1$ and $\beta_2$ equally well.

IBS investigation support the previous results, with the true model having the highest and the baseline model the lowest IBS. Table~\ref{tab:simulation} reports the performance of DeepPAM and the PAM baseline model relative to the performance of the ground truth model at the three quartiles of the survival time distribution. 
In addition, we also include the Kaplan-Meier estimator and the Cox PH model in the comparison. 
DeepPAM outperforms all models (except for the correctly specified PAM).

\begin{table}[!htbp] \centering 
\begin{tabular}{@{\extracolsep{3pt}} cccc} 
\\[-1.8ex]\hline 
\hline \\[-1.8ex] 
 & 1st Quartile & Median & 3rd Quartile \\ 
\hline \\[-1.8ex] 
KM & $6.65$ $(3.25)$ & $12.54$ $(5.47)$ & $19.11$ $(6.87)$ \\ 
Cox PH & $2.18$ $(1.60)$ & $4.16$ $(3.03)$ & $7.03$ $(4.07)$ \\ 
PAM & $2.43$ $(1.79)$ & $4.39(3.13)$ & $7.22$ $(3.97)$ \\ 
DeepPAM & $\mathbf{0.97}$ $(1.39)$ & $\mathbf{1.27}$ $(1.83)$ & $\mathbf{2.41}$ $(2.06)$ \\ 
\hline \\[-1.8ex] 
\end{tabular} 
\caption{Averaged relative difference of the IBS (in \%, standard deviation in brackets), compared to the correct PAM, for different quantiles (columns) and the different models (rows). 
We report the Kaplan-Meier estimator (KM), a Cox PH and a PAM model without point cloud information (Cox PH / PAM).
The closer values are to the value 0, the more similar the performance of the model to the correct PAM.} 
\label{tab:simulation} 
\end{table}

\section{Outlook}

We present DeepPAM, a novel framework for survival analysis to seamlessly combine different data modalities.
While the framework is very general, we demonstrate the use of DeepPAM to jointly model point clouds and tabular data. 
We show that DeepPAM yields comparable results to other benchmark models on an Alzheimer's disease prediction task. In this example, incorporating unstructured information in the model does not seem to improve predictions notably, but the DeepPAM yields slightly better results than the wide-and-deep approach by \citet{polsterl2019wide}. We further demonstrate the efficacy of DeepPAM on a synthetic data set, where additional unstructured data information improves the model's predictions. 

For future research, DeepPAM needs to be tested in more complex settings and applications to reinforce its value to time-to-event analysis. 
As the presented framework is flexible, various combinations of structured PAMs with (deep) neural networks are possible, in particular, deviations from the PH assumption and application to more complex survival tasks. 

\pagebreak

\section*{Acknowledgments}
This work has been partly funded by the Bavarian State Ministry of Education,
Science and the Arts (SP, CW) in the framework of the Centre Digitisation.Bavaria (ZD.B) and the German Federal Ministry of Education and Research (BMBF) under Grant Nos. 01IS18036A (PK, AB, BB, DR) and 031L0200A (SP, CW). The authors of this work take full responsibilities for its content.

\bibliography{preprint}

\begin{thebibliography}{38}
\providecommand{\natexlab}[1]{#1}
\providecommand{\url}[1]{\texttt{#1}}
\providecommand{\urlprefix}{URL }
\expandafter\ifx\csname urlstyle\endcsname\relax
  \providecommand{\doi}[1]{doi:\discretionary{}{}{}#1}\else
  \providecommand{\doi}{doi:\discretionary{}{}{}\begingroup
  \urlstyle{rm}\Url}\fi

\bibitem[{Alaa and {van der Schaar}(2017)}]{alaa_2017}
Alaa, A.~M.; and {van der Schaar}, M. 2017.
\newblock Deep multi-task gaussian processes for survival analysis with
  competing risks.
\newblock In \emph{Proceedings of the 31st International Conference on Neural
  Information Processing Systems}, 2326--2334. Curran Associates Inc.

\bibitem[{Argyropoulos and Unruh(2015)}]{argyropoulos_analysis_2015}
Argyropoulos, C.; and Unruh, M.~L. 2015.
\newblock Analysis of {Time} to {Event} {Outcomes} in {Randomized} {Controlled}
  {Trials} by {Generalized} {Additive} {Models}.
\newblock \emph{PLoS ONE} 10(4): e0123784.

\bibitem[{Baumann, Hothorn, and R{\"u}gamer(2020)}]{baumann2020deep}
Baumann, P.; Hothorn, T.; and R{\"u}gamer, D. 2020.
\newblock Deep Conditional Transformation Models.
\newblock \emph{arXiv preprint arXiv:2010.07860} .

\bibitem[{Bender, Groll, and Scheipl(2018)}]{Bender2018b}
Bender, A.; Groll, A.; and Scheipl, F. 2018.
\newblock A generalized additive model approach to time-to-event analysis.
\newblock \emph{Statistical Modelling} 1471082X17748083.

\bibitem[{Bender et~al.(2020)Bender, R{\"u}gamer, Scheipl, and
  Bischl}]{bender2020general}
Bender, A.; R{\"u}gamer, D.; Scheipl, F.; and Bischl, B. 2020.
\newblock A General Machine Learning Framework for Survival Analysis.
\newblock \emph{arXiv preprint arXiv:2006.15442} .

\bibitem[{Bender and Scheipl(2018)}]{pammtools}
Bender, A.; and Scheipl, F. 2018.
\newblock pammtools: {Piece}-wise exponential {Additive} {Mixed} {Modeling}
  tools.
\newblock \emph{arXiv:1806.01042 [stat]} ArXiv: 1806.01042.

\bibitem[{Bender et~al.(2019)Bender, Scheipl, Hartl, Day, and
  K\"uchenhoff}]{Bender2018a}
Bender, A.; Scheipl, F.; Hartl, W.; Day, A.~G.; and K\"uchenhoff, H. 2019.
\newblock Penalized estimation of complex, non-linear exposure-lag-response
  associations.
\newblock \emph{Biostatistics} 20(2): 315--331.

\bibitem[{Biganzoli, Boracchi, and Marubini(2002)}]{biganzoli_general_2002}
Biganzoli, E.; Boracchi, P.; and Marubini, E. 2002.
\newblock A general framework for neural network models on censored survival
  data.
\newblock \emph{Neural Networks} 15(2): 209--218.

\bibitem[{Cai, Hyndman, and Wand(2002)}]{Cai2002}
Cai, T.; Hyndman, R.~J.; and Wand, M.~P. 2002.
\newblock Mixed Model-Based Hazard Estimation.
\newblock \emph{Journal of Computational and Graphical Statistics} 11(4):
  784--798.

\bibitem[{Cates et~al.(2008)Cates, Fletcher, Styner, Hazlett, and
  Whitaker}]{cates2008particle}
Cates, J.; Fletcher, P.~T.; Styner, M.; Hazlett, H.~C.; and Whitaker, R. 2008.
\newblock Particle-based shape analysis of multi-object complexes.
\newblock In \emph{International Conference on Medical Image Computing and
  Computer-Assisted Intervention}, 477--485. Springer.

\bibitem[{Faraggi and Simon(1995)}]{faraggi_neural_1995}
Faraggi, D.; and Simon, R. 1995.
\newblock A neural network model for survival data.
\newblock \emph{Statistics in Medicine} 14(1): 73--82.

\bibitem[{Fischl(2012)}]{fischl2012freesurfer}
Fischl, B. 2012.
\newblock FreeSurfer.
\newblock \emph{Neuroimage} 62(2): 774--781.

\bibitem[{Fornili et~al.(2014)Fornili, Ambrogi, Boracchi, and
  Biganzoli}]{fornili_piecewise_2014}
Fornili, M.; Ambrogi, F.; Boracchi, P.; and Biganzoli, E. 2014.
\newblock Piecewise {Exponential} {Artificial} {Neural} {Networks} ({PEANN})
  for {Modeling} {Hazard} {Function} with {Right} {Censored} {Data}.
\newblock In Formenti, E.; Tagliaferri, R.; and Wit, E., eds.,
  \emph{Computational {Intelligence} {Methods} for {Bioinformatics} and
  {Biostatistics}}, Lecture {Notes} in {Computer} {Science}, 125--136. Cham:
  Springer International Publishing.

\bibitem[{Friedman(1982)}]{Friedman1982}
Friedman, M. 1982.
\newblock Piecewise Exponential Models for Survival Data with Covariates.
\newblock \emph{The Annals of Statistics} 10(1): 101--113.

\bibitem[{Frisoni et~al.(2008)Frisoni, Ganzola, Canu, R{\"u}b, Pizzini,
  Alessandrini, Zoccatelli, Beltramello, Caltagirone, and
  Thompson}]{frisoni2008mapping}
Frisoni, G.~B.; Ganzola, R.; Canu, E.; R{\"u}b, U.; Pizzini, F.~B.;
  Alessandrini, F.; Zoccatelli, G.; Beltramello, A.; Caltagirone, C.; and
  Thompson, P.~M. 2008.
\newblock Mapping local hippocampal changes in Alzheimer's disease and normal
  ageing with MRI at 3 Tesla.
\newblock \emph{Brain} 131(12): 3266--3276.

\bibitem[{Gerardin et~al.(2009)Gerardin, Ch{\'e}telat, Chupin, Cuingnet,
  Desgranges, Kim, Niethammer, Dubois, Leh{\'e}ricy, Garnero
  et~al.}]{gerardin2009multidimensional}
Gerardin, E.; Ch{\'e}telat, G.; Chupin, M.; Cuingnet, R.; Desgranges, B.; Kim,
  H.-S.; Niethammer, M.; Dubois, B.; Leh{\'e}ricy, S.; Garnero, L.; et~al.
  2009.
\newblock Multidimensional classification of hippocampal shape features
  discriminates Alzheimer's disease and mild cognitive impairment from normal
  aging.
\newblock \emph{Neuroimage} 47(4): 1476--1486.

\bibitem[{Graf et~al.(1999)Graf, Schmoor, Sauerbrei, and
  Schumacher}]{graf1999assessment}
Graf, E.; Schmoor, C.; Sauerbrei, W.; and Schumacher, M. 1999.
\newblock Assessment and comparison of prognostic classification schemes for
  survival data.
\newblock \emph{Statistics in medicine} 18(17-18): 2529--2545.

\bibitem[{Greven and Scheipl(2020)}]{greven_comments_2020}
Greven, S.; and Scheipl, F. 2020.
\newblock Comments on: {Inference} and computation with {Generalized}
  {Additive} {Models} and their extensions.
\newblock \emph{TEST} ISSN 1863-8260.
\newblock \doi{10.1007/s11749-020-00714-2}.
\newblock \urlprefix\url{https://doi.org/10.1007/s11749-020-00714-2}.

\bibitem[{Holford(1980)}]{Holford1980}
Holford, T.~R. 1980.
\newblock The Analysis of Rates and of Survivorship Using Log-Linear Models.
\newblock \emph{Biometrics} 36(2): 299--305.

\bibitem[{Jack~Jr et~al.(2008)Jack~Jr, Bernstein, Fox, Thompson, Alexander,
  Harvey, Borowski, Britson, L.~Whitwell, Ward et~al.}]{jack2008alzheimer}
Jack~Jr, C.~R.; Bernstein, M.~A.; Fox, N.~C.; Thompson, P.; Alexander, G.;
  Harvey, D.; Borowski, B.; Britson, P.~J.; L.~Whitwell, J.; Ward, C.; et~al.
  2008.
\newblock The Alzheimer's disease neuroimaging initiative (ADNI): MRI methods.
\newblock \emph{Journal of Magnetic Resonance Imaging: An Official Journal of
  the International Society for Magnetic Resonance in Medicine} 27(4):
  685--691.

\bibitem[{Kauermann(2005)}]{kauermann_penalized_2005}
Kauermann, G. 2005.
\newblock Penalized spline smoothing in multivariable survival models with
  varying coefficients.
\newblock \emph{Computational Statistics \& Data Analysis} 49(1): 169--186.

\bibitem[{Kingma and Ba(2015)}]{kingma2015adam}
Kingma, D.~P.; and Ba, J. 2015.
\newblock Adam: {A} Method for Stochastic Optimization.
\newblock In Bengio, Y.; and LeCun, Y., eds., \emph{3rd International
  Conference on Learning Representations, {ICLR} 2015, San Diego, CA, USA, May
  7-9, 2015, Conference Track Proceedings}.
\newblock \urlprefix\url{http://arxiv.org/abs/1412.6980}.

\bibitem[{Kvamme and Borgan(2019)}]{kvamme_continuous_2019}
Kvamme, H.; and Borgan, O. 2019.
\newblock Continuous and {Discrete}-{Time} {Survival} {Prediction} with
  {Neural} {Networks}.
\newblock \emph{arXiv preprint arXiv:1910.06724} .

\bibitem[{Laird and Olivier(1981)}]{Laird1981}
Laird, N.; and Olivier, D. 1981.
\newblock Covariance Analysis of Censored Survival Data Using Log-Linear
  Analysis Techniques.
\newblock \emph{Journal of the American Statistical Association} 76(374):
  231--240.

\bibitem[{Langa and Levine(2014)}]{langa2014diagnosis}
Langa, K.~M.; and Levine, D.~A. 2014.
\newblock The diagnosis and management of mild cognitive impairment: a clinical
  review.
\newblock \emph{Jama} 312(23): 2551--2561.

\bibitem[{Lee, Yoon, and {van der Schaar}(2020)}]{lee_dynamic-deephit_2020}
Lee, C.; Yoon, J.; and {van der Schaar}, M. 2020.
\newblock Dynamic-{DeepHit}: {A} {Deep} {Learning} {Approach} for {Dynamic}
  {Survival} {Analysis} {With} {Competing} {Risks} {Based} on {Longitudinal}
  {Data}.
\newblock \emph{IEEE transactions on bio-medical engineering} 67(1): 122--133.

\bibitem[{Lee et~al.(2018)Lee, Zame, Yoon, and {van der
  Schaar}}]{lee_deephit_2018}
Lee, C.; Zame, W.~R.; Yoon, J.; and {van der Schaar}, M. 2018.
\newblock {DeepHit}: {A} {Deep} {Learning} {Approach} to {Survival} {Analysis}
  {With} {Competing} {Risks}.
\newblock In \emph{Thirty-{Second} {AAAI} {Conference} on {Artificial}
  {Intelligence}}.

\bibitem[{Liest{\o}l, Andersen, and Andersen(1994)}]{liestbl_survival_1994}
Liest{\o}l, K.; Andersen, P.~K.; and Andersen, U. 1994.
\newblock Survival analysis and neural nets.
\newblock \emph{Statistics in Medicine} 13(12): 1189--1200.

\bibitem[{Petersen(2011)}]{petersen2011mild}
Petersen, R.~C. 2011.
\newblock Mild cognitive impairment.
\newblock \emph{New England Journal of Medicine} 364(23): 2227--2234.

\bibitem[{P{\"o}lsterl et~al.(2019)P{\"o}lsterl, Sarasua, Guti{\'e}rrez-Becker,
  and Wachinger}]{polsterl2019wide}
P{\"o}lsterl, S.; Sarasua, I.; Guti{\'e}rrez-Becker, B.; and Wachinger, C.
  2019.
\newblock A Wide and Deep Neural Network for Survival Analysis from Anatomical
  Shape and Tabular Clinical Data.
\newblock In \emph{Joint European Conference on Machine Learning and Knowledge
  Discovery in Databases}, 453--464. Springer.

\bibitem[{Qi et~al.(2017)Qi, Su, Mo, and Guibas}]{qi2017pointnet}
Qi, C.~R.; Su, H.; Mo, K.; and Guibas, L.~J. 2017.
\newblock Pointnet: Deep learning on point sets for 3d classification and
  segmentation.
\newblock In \emph{Proceedings of the IEEE conference on computer vision and
  pattern recognition}, 652--660.

\bibitem[{Ranganath et~al.(2016)Ranganath, Perotte, Elhadad, and
  Blei}]{ranganath_deep_2016}
Ranganath, R.; Perotte, A.; Elhadad, N.; and Blei, D. 2016.
\newblock Deep {Survival} {Analysis}.
\newblock In \emph{Proceedings of the 1st Machine Learning for Healthcare
  Conference}, volume~59, 101--114.

\bibitem[{R{\"u}gamer, Kolb, and Klein(2020)}]{rugamer_semistructured_2021}
R{\"u}gamer, D.; Kolb, C.; and Klein, N. 2020.
\newblock Semi-Structured Deep Distributional Regression: Combining Structured
  Additive Models and Deep Learning.
\newblock \emph{arXiv preprint arXiv:2002.05777} .

\bibitem[{R{\"u}gamer, Pfisterer, and Bischl(2020)}]{ruegamer2020neural}
R{\"u}gamer, D.; Pfisterer, F.; and Bischl, B. 2020.
\newblock Neural Mixture Distributional Regression.
\newblock \emph{arXiv preprint arXiv:2010.06889} .

\bibitem[{Wachinger et~al.(2016)Wachinger, Salat, Weiner, Reuter, and
  Initiative}]{wachinger2016whole}
Wachinger, C.; Salat, D.~H.; Weiner, M.; Reuter, M.; and Initiative, A. D.~N.
  2016.
\newblock Whole-brain analysis reveals increased neuroanatomical asymmetries in
  dementia for hippocampus and amygdala.
\newblock \emph{Brain} 139(12): 3253--3266.

\bibitem[{Whitehead(1980)}]{Whitehead1980}
Whitehead, J. 1980.
\newblock Fitting Cox's Regression Model to Survival Data using GLIM.
\newblock \emph{Journal of the Royal Statistical Society. Series C (Applied
  Statistics)} 29(3): 268--275.

\bibitem[{Wood(2017)}]{wood_generalized_2017}
Wood, S.~N. 2017.
\newblock \emph{Generalized {Additive} {Models}: {An} {Introduction} with {R}}.
\newblock Boca Raton: Chapman \& Hall/Crc Texts in Statistical Science, 2 rev
  ed. edition.

\bibitem[{Wu et~al.(2015)Wu, Song, Khosla, Yu, Zhang, Tang, and
  Xiao}]{wu20153d}
Wu, Z.; Song, S.; Khosla, A.; Yu, F.; Zhang, L.; Tang, X.; and Xiao, J. 2015.
\newblock 3d shapenets: A deep representation for volumetric shapes.
\newblock In \emph{Proceedings of the IEEE conference on computer vision and
  pattern recognition}, 1912--1920.

\end{thebibliography}

\end{document}